\pdfoutput=1

\documentclass[10pt,twocolumn,letterpaper]{article}

\usepackage{amsmath}

\DeclareMathOperator*{\argmin}{arg\,min}
\usepackage{pifont}
\newcommand{\cmark}{\ding{51}}%
\newcommand{\xmark}{\ding{55}}%
\usepackage{multirow}
\usepackage{xcolor}
\newcommand{\gray}[1]{\textcolor{gray}{#1}}
\usepackage{array, makecell} %
\usepackage{float}

\usepackage{cvpr}              

\usepackage{graphicx}
\usepackage{amsmath}
\usepackage{amssymb}
\usepackage{booktabs}
\usepackage[accsupp]{axessibility}  

\usepackage[pagebackref,breaklinks,colorlinks]{hyperref}

\usepackage[capitalize]{cleveref}
\crefname{section}{Sec.}{Secs.}
\Crefname{section}{Section}{Sections}
\Crefname{table}{Table}{Tables}
\crefname{table}{Tab.}{Tabs.}

\def\cvprPaperID{2303} 
\def\confName{CVPR}
\def\confYear{2023}

\begin{document}

\title{Catch Missing Details: Image Reconstruction with Frequency Augmented Variational Autoencoder}

\author{Xinmiao Lin\\
Rochester Institute of Technology\\
{\tt\small xl3439@rit.edu}
\and
Yikang Li\\
OPPO US Research\\
{\tt\small yikang.li1@oppo.com}
\and
Jenhao Hsiao\\
OPPO US Research\\
{\tt\small mark@oppo.com}
\and
Chiuman Ho\\
OPPO US Research\\
{\tt\small chiuman@oppo.com}
\and
Yu Kong\\
Michigan State University\\
{\tt\small yukong@msu.edu}
}

\maketitle

\begin{abstract}
    The popular VQ-VAE models reconstruct images through learning a discrete codebook but suffer from a significant issue in the rapid quality degradation of image reconstruction as the compression rate rises. One major reason is that a higher compression rate induces more loss of visual signals on the higher frequency spectrum which reflect the details on pixel space. In this paper, a Frequency Complement Module (FCM) architecture is proposed to capture the missing frequency information for enhancing reconstruction quality. The FCM can be easily incorporated into the VQ-VAE structure, and we refer to the new model as \textbf{F}requancy \textbf{A}ugmented \textbf{VAE} (\textbf{FA-VAE}). In addition, a Dynamic Spectrum Loss (DSL) is introduced to guide the FCMs to balance between various frequencies dynamically for optimal reconstruction. FA-VAE is further extended to the text-to-image synthesis task, and a Cross-attention Autoregressive Transformer (CAT) is proposed to obtain more precise semantic attributes in texts. Extensive reconstruction experiments with different compression rates are conducted on several benchmark datasets, and the results demonstrate that the proposed FA-VAE is able to restore more faithfully the details compared to SOTA methods. CAT also shows improved generation quality with better image-text semantic alignment. Code is available at \url{https://github.com/oppo-us-research/FA-VAE}

\end{abstract}

\section{Introduction}
\label{sec:intro}

\begin{figure}[t!]
    \centerline{\includegraphics[width=0.48\textwidth]{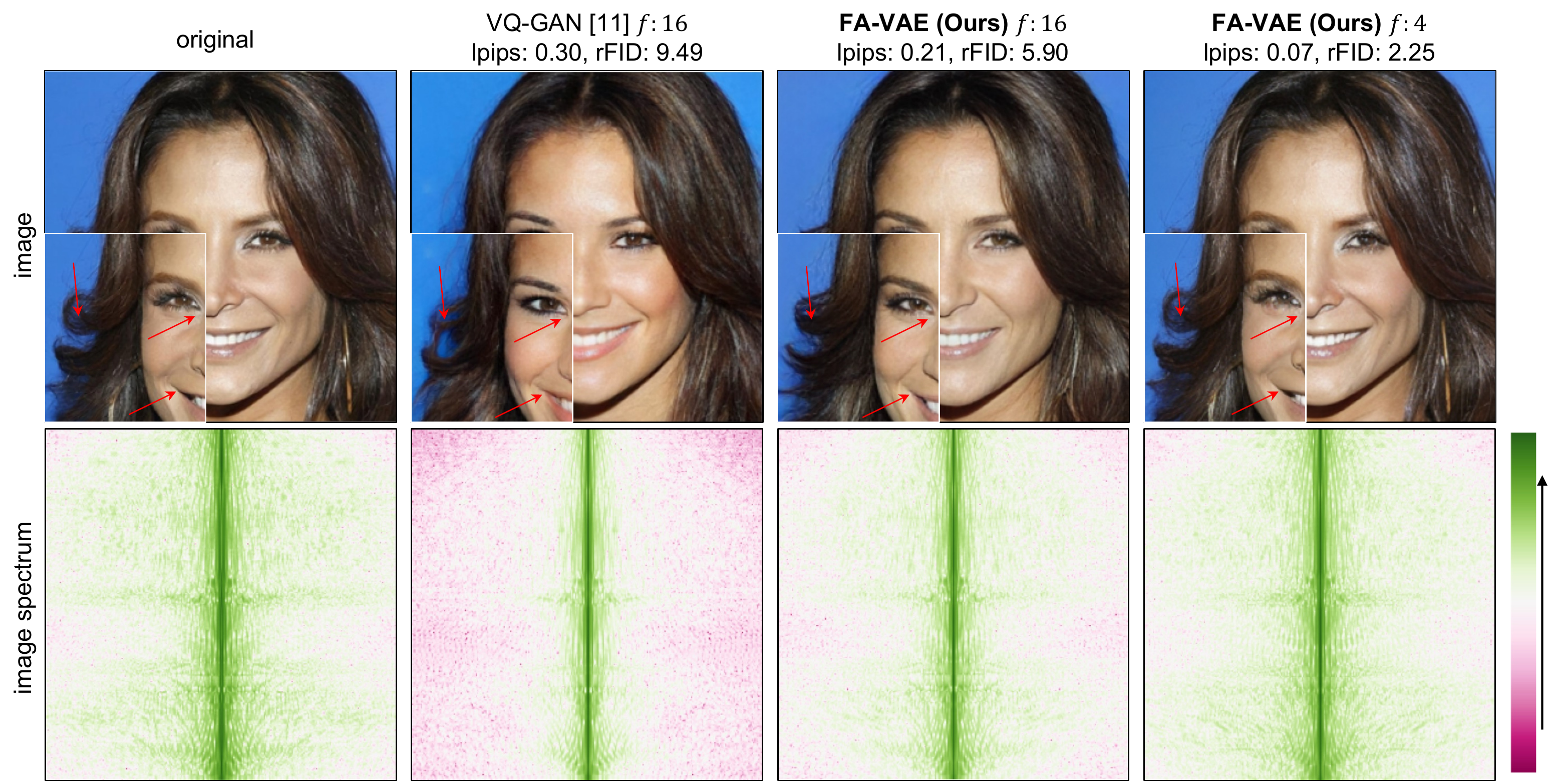}}
    \captionsetup{aboveskip=3pt}
    \caption{\small {Images and their frequency maps. Row 1: original and reconstructed images. Row 2: the frequency maps of images, frequency increases in any direction away from the center. $f$ is the compression rate. With a greater compression rate, more details are lost during reconstruction, i.e. eyes and mouth shape, and hair texture (pointed with red arrows) which align with the loss of high-frequency features. All frequency figures in this paper use the same colormap. rFID \cite{fid} and lpips \cite{lpips} are lower the better, and frequency values increase from red to green, zoom in for better visualization. }
    }
    \label{fig:first}
    \vspace{-0.7cm}
\end{figure}

VQ-VAE models \cite{vq-vae, vqvae-2, dc-vae, deepVAE, taming, rq-vae} reconstruct images through learning a discrete codebook of latent embeddings. They gained wide popularity due to the scalable and versatile codebook, which can be broadly applied to many visual tasks such as image synthesis \cite{clip-gen, taming} and inpainting \cite{inpainting-hvqvae, maskgit}. A higher compression rate is typically preferable in VQ-VAE models since it provides memory efficiency and better learning of coherent semantics structures \cite{latent-diffusion, taming}.

One main challenge quickly arises for a higher compression rate, which severely compromises reconstruction accuracy. Figure \ref{fig:first} row 1 shows that although the reconstructed images at higher compression rates appear consistent with the original image, details inconsistencies such as the color and contour of the lips become apparent upon closer scrutiny. Figure \ref{fig:first} row 2 reveals that similar degradation also manifests on the frequency domains where features towards the middle and higher frequency spectrum are the least recoverable with greater compression rate. 

Several causes stand behind this gap between pixel and frequency space. The convolutional nature of autoencoders is prone to \textit{spectral bias}, which favors learning low-frequency features \cite{zoomtoinpaint, spectral-bias}. This challenge is further aggravated when current methods exclusively design losses or improve model architecture for better semantics resemblance \cite{taming, lpips, rq-vae} but often neglect the alignment on the frequency domain \cite{ambp, ffl}. On top of that, it is intuitively more challenging for a decoder to reconstruct an image patch from a single codebook embedding (high compression) than multiple embeddings (less compression). The reason is that the former mixes up features of incomplete and diverse frequencies, while the latter could preserve more fine-grained and complete features at various frequencies.

Inspired by these insights, the \textit{\textbf{F}requency \textbf{A}ugmented \textbf{VAE} (\textbf{FA-VAE})} model is proposed, which aims to improve reconstruction quality by achieving better alignment on the frequency spectrums between the original and reconstructed images. More specifically, new modules named \textit{Frequency Complement Modules (FCM)} are crafted and embedded at multiple layers of FA-VAE's decoder to learn to complement the decoder's features with missing frequencies. 

We observe that valuable middle and high frequencies are mingled with the encoder's feature maps during the compression via an encoder, shown in Figure \ref{fig:method_freq} row 4. Therefore, a new loss termed \textit{Spectrum Loss (SL)} is proposed to guide FCMs to generate missing features that align with the same level's encoder features on the frequency domain. Since most image semantics reside on the low-frequency spectrum \cite{frequency-cnn}, SL prioritizes learning lower-frequency features with diminishing weights as frequencies increase.  

Interestingly, we discover that checkerboard patterns appear in the complemented decoder's features with SL, although better reconstruction performance is achieved (Figure \ref{fig:method_freq} column 4). We speculate that because SL sets a deterministic range for the low-frequency spectrum when applying weights on the frequencies without considering that the importance of a frequency can vary from layer to layer. Thus, an improved loss function \textit{Dynamic Spectrum Loss (DSL)} is crafted on top of SL with a learnable component to adjust the range of low-frequency spectrum dynamically for optimal reconstruction. DSL can improve reconstruction quality even further than SL without the unnatural checkerboard artifacts in the features (Figure \ref{fig:method_freq} column 5).

We further extend FA-VAE to the text-to-image generation task and propose the \textit{Cross-attention Autoregressive Transformer (CAT)} model. We first observe that only using one or a few token embeddings is a coarse representation of lengthy texts \cite{clip, vqganclip, glide}. Thus CAT uses all token embeddings as a condition for more precise guidance. Moreover, existing works typically use self-attention, and the text condition is embedded merely at the beginning of the generation \cite{clip-gen, taming}. This mechanism becomes problematic in the autoregressive generation because one image token is generated at a time, thus the text condition gradually loosens its connection with the generated tokens. To circumvent this issue, CAT embeds a cross-attention mechanism that allows the text condition to guide each step generation.

To summarize, our work includes the following contributions:
\begin{itemize}
    \vspace{-0.3cm}
    \item We propose a new type of architecture called \textit{Frequency Augmented VAE (FA-VAE)} for improving image reconstruction through achieving more accurate details reconstruction. 
    \vspace{-0.3cm}
    \item We propose a new loss called \textit{Spectrum Loss (SL)} and its enhanced version \textit{Dynamic Spectrum Loss (D SL)}, which guides the \textit{Frequency Complement Modules (FCM)} in FA-VAE to adaptively learn different low/high frequency mixtures for optimal reconstruction.
    \vspace{-0.3cm}
    \item We propose a new \textit{Cross-attention Autoregressive Transformer (CAT)} for text-to-image generation using more fine-grained textual embeddings as a condition with a cross-attention mechanism for better image-text semantic alignment.
\end{itemize}

\section{Related Work \& Background}
\label{sec:background}

\begin{figure*}[t!]
    \centerline{\includegraphics[width=1.\textwidth]{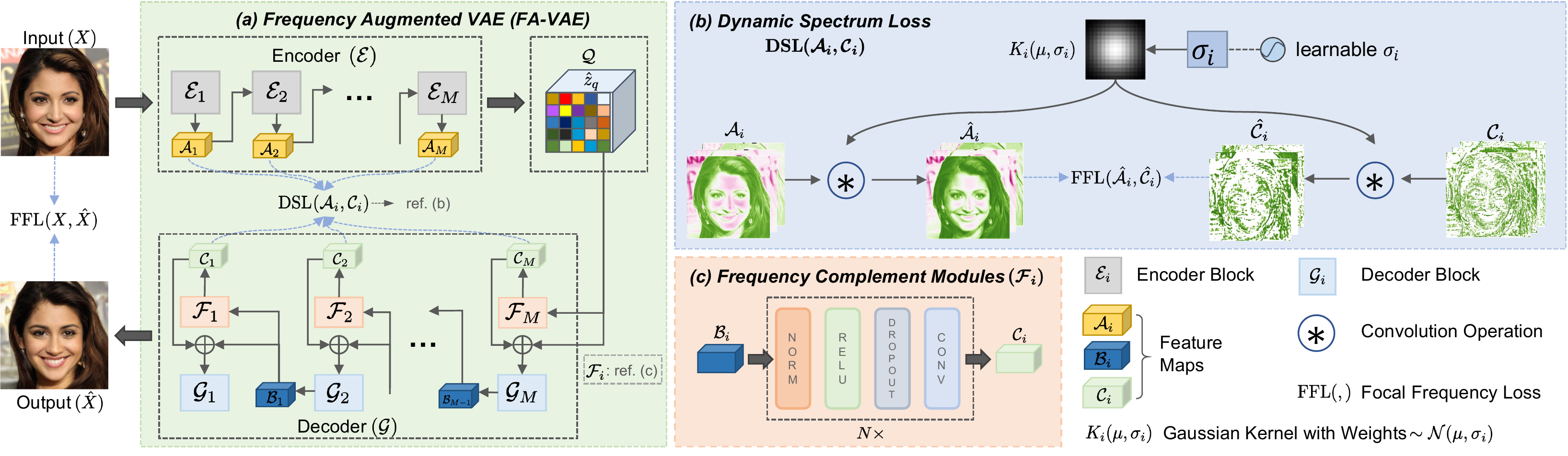}}
    \captionsetup{aboveskip=3pt}
    \caption{
    \small{\textit{\textbf{F}requency \textbf{A}ugmented VAE (\textbf{FA-VAE})}}. (a) The encoder $\mathcal{E}$ encodes the images $X$ onto discrete latent codebook space $\hat{z}_q$ which is used by the decoder $\mathcal{G}$ to reconstruct images $\hat{X}$. DSL (b) guides the FCMs (c) to learn to complement the reconstructed features with missing features of important frequencies in order to improve reconstruction quality. A more detailed figure is in the supplement. }
    \label{fig:overview}
    \vspace{-0.6cm}
\end{figure*}

\textbf{Image Reconstruction} Vector Quantized-Variational AutoEncoder (VQ-VAE) \cite{vq-vae} extends the Variational AutoEncoder structure \cite{vae} and proposes to encode images into discrete latent codes using vector quantization (VQ). Then a generative model, such as an autoregressive transformer \cite{autoregressive-net}, can be trained and paired with the decoder in VQ-VAE to synthesize new images. Later works \cite{vqvae-2, taming, rq-vae} further improve the reconstruction and generation quality by improving the generative model architecture or the quantization efficiency. Since VQ-VAE models operate on discrete latent spaces, they cannot be directly and fairly compared to other VAE-based models that employ continuous latent space \cite{vae, vaebm, nvae, dc-vae, deepVAE, ncp-vae}. In contrast to current VQ-VAE models, our proposed FA-VAE technique improves reconstruction through the frequency angle and is more generalized since the FCMs can be readily extended to other neural networks that share the same VQ-VAE structure.


Few other works do image reconstruction via the frequency perspective. FFL \cite{ffl} proposed a loss function to penalize differences on the hard frequencies. A new module, which is designed in the continuous latent space for image compression, is proposed to be incorporated into the encoder and decoder in \cite{ambp}. To our best knowledge, FA-VAE is the first work that aims to improve image reconstruction on discrete latent space through the frequency perspective.

\textbf{Image Generation} Image generation can be achieved via GAN-based models \cite{progressive-gan, unet-gan, large-gan, dmgan, attngan, dfgan}, which synthesize images from noise vectors unconditionally or conditioned on different inputs such as texts, masks, etc. It becomes cumbersome to train one model for one application. Thus, StyleGAN \cite{ffhq} encodes semantic attributes into a continuous latent space, and subsequent works \cite{stylegan2-ada, stylegan3, stylealae, lafite, tedigan, stylemc, anyface} leverage this space for generating images conditioned on attributes or textual descriptions. However, StyleGAN-based models cannot scale to large datasets when the number of attributes becomes substantially large because this demands to increase model's size. In contrast, the codebook in VQ-VAE models \cite{vq-vae} is scalable to large datasets without additional model complexity. Diffusion-based models \cite{dalle2, latent-diffusion, glide, chen2022analog, Choi2022PerceptionPT, imagen, ddim, ddpm} generate images from Gaussian noise through a reverse diffusion process can often require substantial training and large datasets. Most text-to-image generation models commonly use one of a few embeddings for a condition \cite{dalle, clip-gen, glide, vqganclip} from pretrained models such as CLIP \cite{clip} or T5-XXL \cite{t5-xxl}. In contrast, our proposed autoregressive transformer CAT uses the embeddings from all the text tokens for more fine-grained guidance during image generation. 

\textbf{VQ-VAE} We now describe the VQ-VAE model \cite{vq-vae} as it is the backbone used in the FA-VAE model, which is presented in Figure \ref{fig:overview}. VQ-VAE \cite{vq-vae} and related models \cite{taming, rq-vae, vqvae-2, dalle} consist of an encoder that encodes the images into a codebook of embeddings. Then, a decoder is trained to reconstruct the images from a discrete set of codebook embeddings. In this paper, reconstruction loss in VQ-GAN is utilized which is: 
\begin{equation}
    \mathcal{L}_{rec} = \lVert {X - \hat{X}} \rVert_1 + \mathcal{L}_{pips} (X-\hat{X}) 
    \label{eq:rec_loss}
\end{equation}
where $X$ and $\hat{X}$ are the original and reconstructed images. We also use the adversarial setting of VQ-GAN model which introduced a discriminator compared to VQ-VAE models \cite{vq-vae, vqvae-2}, more details could be referred to \cite{taming}.

\section{Methodology}
\label{sec:method}

The proposed \textit{\textbf{F}requency \textbf{A}ugmented \textbf{VAE} (\textbf{FA-VAE})} is presented in Figure \ref{fig:overview}. In this section, we describe how FA-VAE ameliorates reconstruction quality by bridging the spectrum domain gap between reconstructed and original images. By explicitly embedding \textit{Frequency Complement Modules (FCM)} into the decoder, \textit{Dynamic Spectrum Loss (DSL)} leverages important features frequency-wise from the encoder and guides the FCMs to complement the features of the decoder for better frequency restoration at different reconstruction stages.  

\subsection{Frequency Augmented VAE (FA-VAE)}
\label{sec:fa-vae-background}
Let the images be $X \in \mathbb{R}^{H \times W \times 3}$. The codebook $\mathfrak{C}$ is a set of $|\mathfrak{C}|$ embeddings, such as $\mathfrak{C} = \{ \mathfrak{c}_i | i = 1, ..., |\mathfrak{C}| \} \in \mathbb{R}^{n_z}$ and $n_z$ is the length of one codebook embedding. FA-VAE consists of an encoder $\mathcal{E}$ that encodes the images $X$ into latent space representations such as $z = \mathcal{E}(x)$, for $z \in \mathbb{R}^{h \times w \times {n_z}}$ and $x \in X$, $(h \times w)$ is the resolution of the encoded representation. Let $f = H/h = W/w$ be the downsampling factor or compression rate. Each feature of $z$ is approximated by the vector quantization block $\mathcal{Q}$ using the nearest codebook entries, and it can be presented as:
\begin{equation}
    \mathcal{Q}(z) = \hat{z}_{q} = \argmin_{c_k \in \mathcal{C}} \lVert z_{ij} - \mathfrak{c}_k \rVert
\end{equation}
where $\hat{z}_{q}$ is the quantized latent embedding and subsequently used by the decoder $\mathcal{G}$ to produce the reconstructed image $\hat{x} = \mathcal{G}(\hat{z}_q)$. 

\subsubsection{Frequency Complement Modules (FCM)}
\label{sec:fcm}
\begin{figure}[t!]
    \centerline{\includegraphics[width=.45\textwidth]{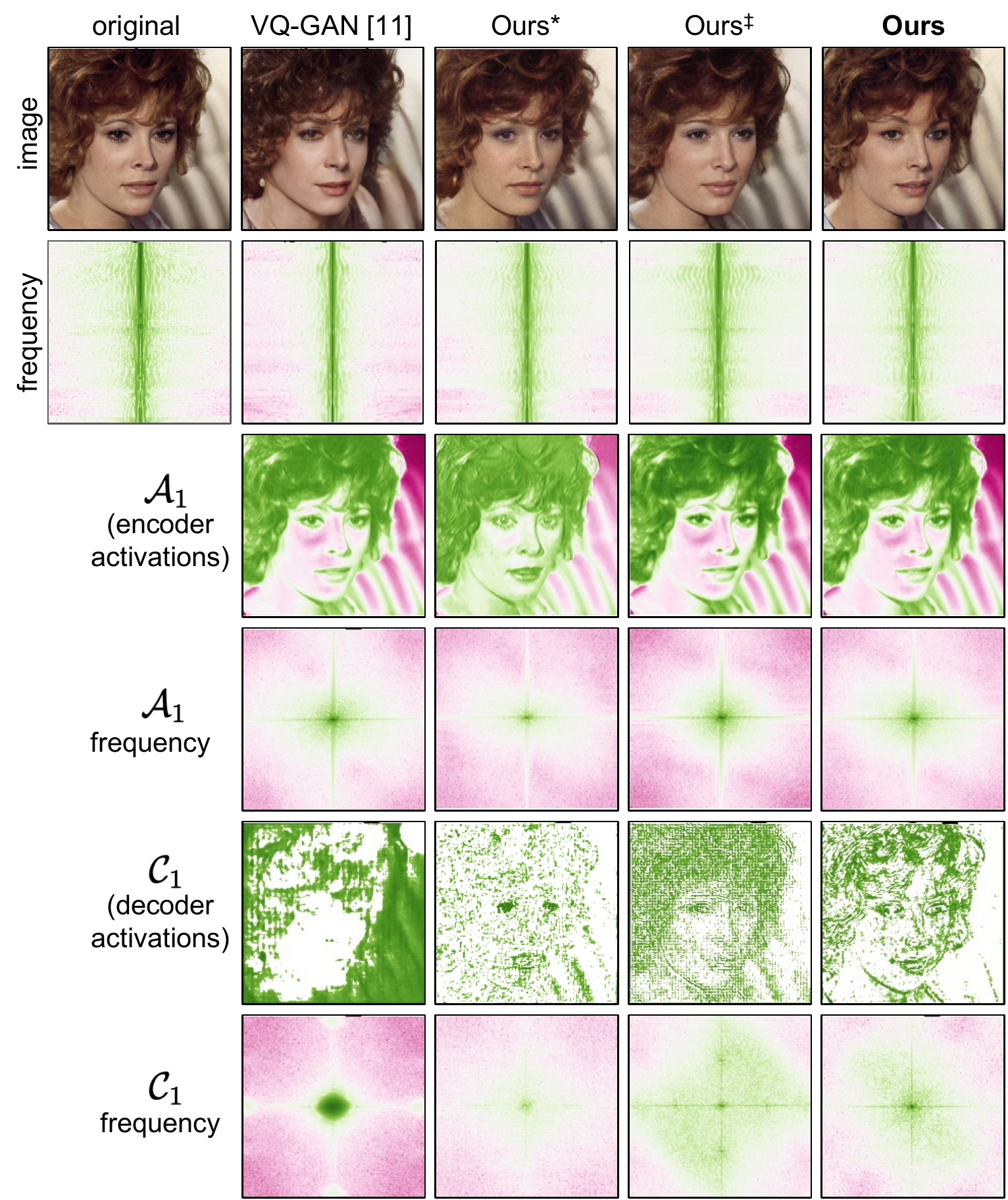}}
    \captionsetup{aboveskip=3pt}
    \caption{Image, activations, and their frequency maps. Ours* is FA-VAE model with FFL \cite{ffl}, $\text{Ours}^{\ddagger}$ is FA-VAE with SL and Ours is FA-VAE model with DSL. DSL shows a more harmonious balance between low- and high-frequencies (row 5) and more accurate reconstruction in mouth and hair textures. The frequency maps are plotted using the average of all channels and the contrast is adjusted to emphasize the higher frequency spectrums. }
    \label{fig:method_freq}
    \vspace{-0.6cm}
\end{figure}

\textbf{Motivation} Figure \ref{fig:first} shows that a higher compression rate leads to more significant reconstruction disparities on the higher-frequency spectrum. Existing models \cite{taming, vqvae-2, rq-vae} guide the reconstructed images to be more aligned on the pixel and feature spaces with the original images (Eq. \ref{eq:rec_loss}) but neglect frequency spectrum alignment and leave the encoder and decoder without further guidance. Figure \ref{fig:method_freq} column 2 shows that the encoder activations $\mathcal{A}_1$ of baseline VQ-GAN \cite{taming} contain rich high-frequency features (row 3 \& 4), but the decoder's activations could mainly restore low-frequency features (row 5 \& 6).

Therefore we propose \textit{Frequency Complement Modules (FCM)}, illustrated in Figure \ref{fig:overview} (c), which aims to complement the decoder's feature maps $\mathcal{B}_i$ with features of missing frequencies using the encoder activations $\mathcal{A}_i$. The FCMs $\mathcal{F}_i$ consist of sequences of convolution layers and activations. The decoder $\mathcal{G}$ with FCMs embedded can be represented as:
\begin{equation}
\begin{aligned}
    \mathcal{B}_{M-1} &= \mathcal{G}_M \left( \mathcal{C}_M + \hat{z}_{q} \right) = \mathcal{G}_M \left(  \mathcal{F}_M( \hat{z}_{q} ) + \hat{z}_{q} \right) \\
    \cdots \\
    \mathcal{B}_1 &= \mathcal{G}_2 (\mathcal{C}_2 + \mathcal{B}_2) = \mathcal{G}_2 ( \mathcal{F}_2(\mathcal{B}_2) + \mathcal{B}_2)
    \end{aligned}
\label{eq:new_generator}
\end{equation}
Similarly, the encoder $\mathcal{E}$ can be abstracted to:
\begin{align}
    \label{eq:encoder}
    z &= \mathcal{E}(x) = \mathcal{E}_{M} \left( \cdots \left( \mathcal{E}_{2} \left( \mathcal{E}_{1}(x) \right) \right) \right) 
\end{align}
where $\mathcal{G}_i$ is the $i$-th layer of the decoder, and $\mathcal{A}_i = \mathcal{E}_i (x)$ is the outputs or feature maps of the corresponding encoder block. The outputs of previous blocks $\mathcal{B}_i$ are complemented by outputs $\mathcal{C}_i = \mathcal{F}_i(\mathcal{B}_{i+1})$, which contain the frequency-rich features learned from the encoder activations $\mathcal{A}_i$. The following section describes how Dynamic Frequency Loss (DSL) guides the learning of FCMs to specifically learn to complement $\mathcal{B}_i$ with the features of missing frequencies.

\begin{figure}[t!]
\begin{subfigure}{.15\textwidth}
  \centering
  \includegraphics[width=.8\linewidth]{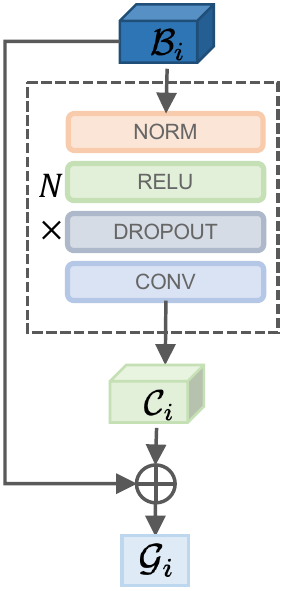}
  \caption{\footnotesize{FCM}}
  \label{fig:sfig1}
\end{subfigure}%
\begin{subfigure}{.15\textwidth}
  \centering
  \includegraphics[width=.8\linewidth]{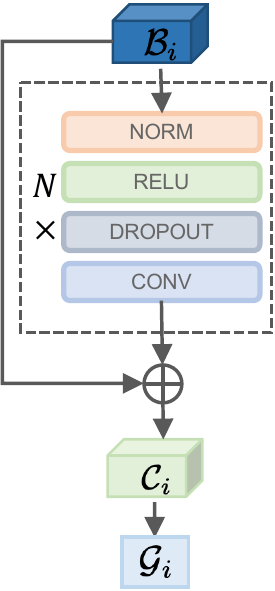}
  \caption{\footnotesize{FCM w/ residual}}
  \label{fig:sfig2}
\end{subfigure}
\begin{subfigure}{.15\textwidth}
  \centering
  \includegraphics[width=.75\linewidth]{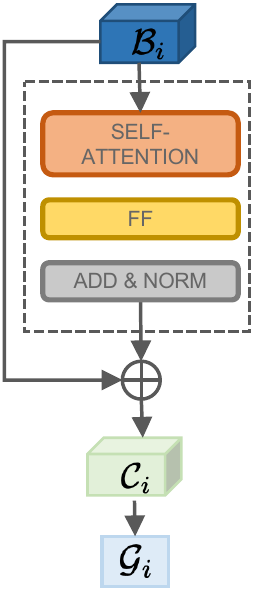}
  \caption{\footnotesize{FCM w/ attention }}
  \label{fig:sfig3}
\end{subfigure}
\captionsetup{aboveskip=3pt}
\caption{FCM with different architectures and connections.}
\label{fig:fcm-architectures}
\vspace{-0.55cm}
\end{figure}

In the paper, the frequency feature compensation is implemented by addition (in Eq. \ref{eq:new_generator}). However, the FCM is flexible in adopting any architecture and merging techniques, as illustrated by the three examples in Figure \ref{fig:fcm-architectures}. Note that no architecture limitation is imposed on the encoder and decoder blocks as long as $\mathcal{A}_i$ and $\mathcal{B}_i$ share the same resolution. VQ-GAN \cite{taming} and related models \cite{rq-vae, vq-vae} already have similar architecture, and VQ-GAN is chosen as the backbone for FA-VAE.

\subsubsection{Dynamic Spectrum Loss (DSL)}
\textbf{Motivation} \textit{Spectrum Loss (SL)} is proposed to guide explicitly the outputs of FCMs $\mathcal{C}_i$ to be more aligned with the encoder's activations $\mathcal{A}_i$ on the frequency spectrum because the latter contains rich features on the higher frequency spectrum (Figure \ref{fig:method_freq}). Moreover, to account for the varying importance of frequencies across different decoder stages, \textit{Dynamic Spectrum Loss (DSL)} is proposed, a more generalized variant of SL, and has the ability to adjust weights put on higher-frequency spectrum adaptively. Thus, each decoder block's outputs can be enriched with features of the most critical frequencies for an accurate reconstruction.


\textbf{Background} The outputs $\mathcal{A}_i$ of encoder block $\mathcal{E}_i$ and $\mathcal{C}_i$ of FCM block $\mathcal{F}_i$ are first transformed to the frequency domain using Discrete Fourier Transform (DFT) as follows:
\begin{equation}
    F(u, v) = \sum_{x=0}^{M-1} \sum_{y=0}^{N-1} f(x, y) \cdot e^{-i2\pi \left(\frac{ux}{M} + \frac{vy}{N} \right)}
\label{eq:dft}
\end{equation}
where $e$ and $i$ are Euler's number and the imaginary unit. $M \times N$ is the spatial resolution of feature maps and the Fourier Transform (Eq. \ref{eq:dft}) is applied to each of them. $f(x, y)$ is the value at $(x,y)$ of each feature map. $F(u, v)$ is the corresponding value at $(u,v)$ coordinates on the frequency spectrum. The Focal Frequency Loss (FFL) \cite{ffl} can be presented as:
\begin{equation}
    \text{FFL}(\mathcal{A}_i, \mathcal{C}_i) = \frac{1}{MN |\mathcal{C}_i|} \sum_{c=0}^{|\mathcal{C}_i|-1} \sum_{u=0}^{M-1} \sum_{v=0}^{N-1} w(u, v) J(u, v) ,
\label{eq:ffl}
\end{equation}
where $w(u, v) = |F_{\mathcal{A}_i} (u, v) - F_{\mathcal{C}_i} (u, v)|$ are the weights put on each frequency. $J(u, v) = |F_{\mathcal{A}_i} (u, v) - F_{\mathcal{C}_i} (u, v)|^2$ is the main error function based on the frequency difference. The $|\mathcal{C}_i|$ is the number of feature maps in $\mathcal{A}_i$ and $\mathcal{C}_i$. Both real and imaginary parts of the frequency domain are considered, more details are in \cite{ffl}. 

\textbf{Limitations of FFL} Eq. \ref{eq:ffl} demonstrates that FFL puts modulating term $w(u, v)$ to focus learning on the hardest frequencies for reconstruction, which are the higher frequencies following our observation (Figures \ref{fig:first} \& \ref{fig:method_freq}). However, this could not be ideal because features on the lower frequencies define the image content, and overemphasizing the higher frequencies could over-constrain the learning and lead to sub-optimal reconstruction, see Figure \ref{fig:celeba-ablation} row 1 column 3 and Table \ref{tab:celeba-ablation} row 3. Moreover, Figure \ref{fig:method_freq} column 3 shows FCMs guided by a simple FFL can improve reconstruction performance ($\text{Ours}^*$). However the frequency maps of the decoder activations $\mathcal{C}_1$ contain excessive noise due to overemphasis on the higher frequency spectrum, and lower frequencies are neglected (row 6). 

\textbf{Spectrum Loss (SL)} Thus, we propose to apply a low-pass filter on the weights $w(u, v)$ in Eq. \ref{eq:ffl} to penalize more mismatch in the lower-frequency domain and gradually diminish the penalizing weights towards higher frequency spectrum. 
Therefore, let the Gaussian kernels be $K_i(\mu, \sigma)$ with weights initialized using mean and standard deviation $(\mu, \sigma_i)$, and applied over the feature maps as:
\begin{equation}
    (\mathcal{\hat{A}}_i, \mathcal{\hat{C}}_i) = (K_i(\mu, \sigma_i) \star \mathcal{A}_i, K_i(\mu, \sigma_i) \star \mathcal{C}_i).
\label{eq:gaussian_activation}
\end{equation}
where the $\star$ is the convolution operation. Then, the \textit{Spectrum Loss (SL)} is defined as:
\begin{equation}
    \text{SL}(\mathcal{A}_i, \mathcal{C}_i) = \text{FFL}(\mathcal{\hat{A}}_i, \mathcal{\hat{C}}_i).
\end{equation}
\textbf{Limitations of SL} Up until now, a fixed Gaussian filter $K_i(\mu, \sigma_i)$ is applied over all $\mathcal{A}_i$ and $\mathcal{C}_i$ of all encoder and decoder blocks. Figure \ref{fig:method_freq} column 4 demonstrates that SL ($\text{Ours}^{\ddagger}$) improves reconstruction on the lower frequency spectrum, leading to better reconstruction than the baseline model (Table \ref{tab:celeba-ablation} row 5 vs row 1). However, checkerboard artifacts are also present (row 5) on $\mathcal{C}_1$. One reason is that deterministic variance $\sigma_i$ assumes that decoder activations across different levels require the same amount of higher frequency features for accurate reconstruction, which could be an over-rigid constraint on the learning. Another reason could be that the same $\sigma_i$ magnifies the checkerboard effects produced by upsampling or striding operations in CNNs which are normally circumvented by subsequent convolutional layers \cite{deep_perceptual, cnn-invariance}. In the observation of the experiments, we find that the checkerboard effects can be intensified with the same $\sigma_i$ in different layers due to the bounding effect between low frequency and high frequency (shown in Figure \ref{fig:method_freq} column 4 row 6).

\begin{figure}[t!]
    \centerline{\includegraphics[width=.48\textwidth]{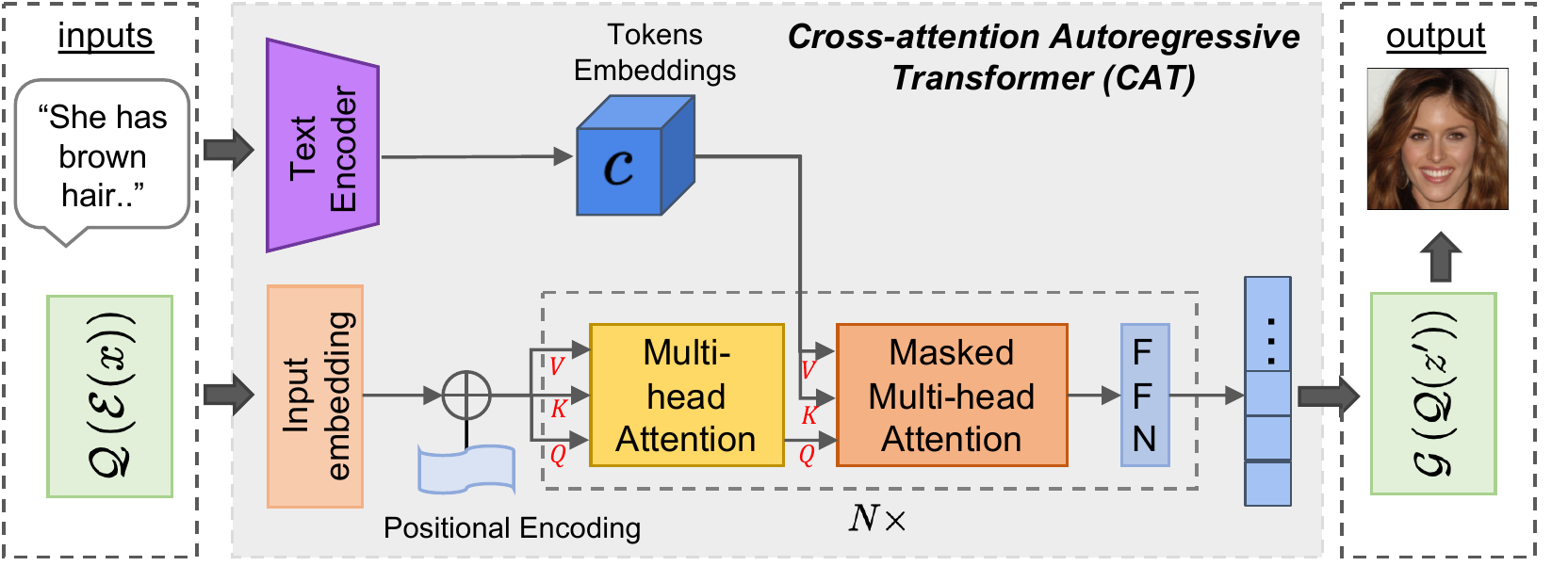}}
    \captionsetup{aboveskip=3pt}
    \caption{\small 
    \small{Cross-attention Autoregressive Transformer (CAT).}}
    \label{fig:prior}
    \vspace{-0.6cm}
\end{figure}

\begin{figure*}[t!]
    \centerline{\includegraphics[width=0.9\textwidth]{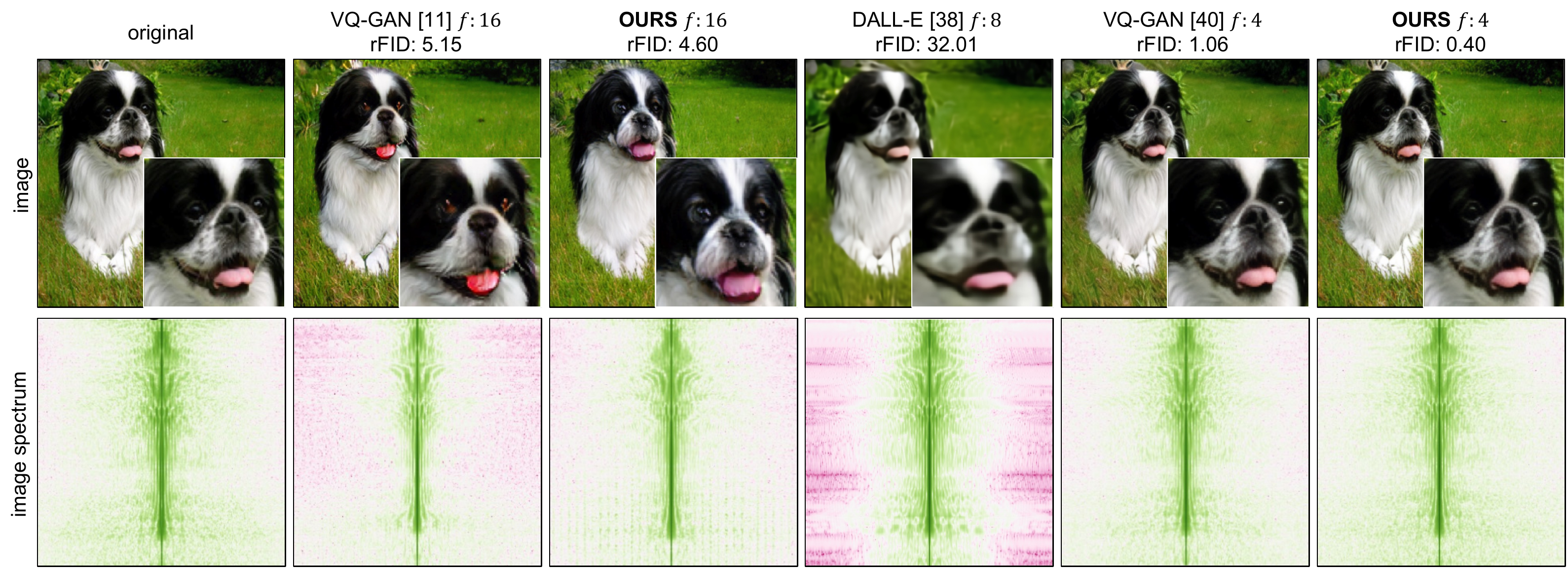}}
    \captionsetup{aboveskip=3pt}
    \caption{\small 
    \small{Reconstruction on ImageNet \cite{imagenet}, label: Japanese terrier.}}
    \label{fig:recon-imagenet}
    \vspace{-0.5cm}
\end{figure*}

\textbf{Dynamic Spectrum Loss (DSL)} Therefore, we propose to optimize the variances $\sigma_i$ instead of setting them as static hyperparameters to suit the different amounts of frequencies needed for each block's $\mathcal{B}_i$. The new \textit{Dynamic Spectrum Loss (DSL)} is a more generalized form of Spectrum Loss (SL) with learnable $\sigma_i$. Note that DSL also includes FFL as a special form when used on the original and reconstructed images without the Gaussian filters.

Then the total reconstruction loss for FA-VAE can be described as:
\begin{equation}
\begin{split}
    \mathcal{L}_{rec} = \alpha \text{FFL}(X, \hat{X}) + \beta \sum_{i=0}^{M-1} \text{DSL}(\mathcal{A}_i, \mathcal{C}_i) \\
                        +\lVert {X - \hat{X}} \rVert_1 + \mathcal{L}_{pips} (X-\hat{X}).
\end{split}
\label{eq:favae-rec}
\end{equation}
where $\text{FFL}(X-\hat{X})$ is the Focal Frequency Loss applied on the original reconstructed images. The second term in Eq. \ref{eq:favae-rec} is the DSL loss applied over the outputs of the encoder and FCM blocks. The first two losses aim to minimize the frequency spectrum differences on the images and internal feature maps. The third and fourth losses act on the pixel and feature maps space \cite{taming}. $\alpha$ and $\beta$ are hyperparameters. 

$\sigma_i$ are model parameters and optimized as:
\begin{equation}
    \sigma_i ^*, \mathcal{E}^*, \mathcal{G}^*, \mathcal{C}^* = \argmin_{\sigma_i, \mathcal{E}, \mathcal{G}, \mathcal{C}} (\mathcal{L}_{rec} + \mathcal{L}_{Q})
\end{equation}
$\mathcal{L}_{Q}$ is the quantization loss which minimizes the difference between the codebook embeddings and the embeddings given by the encoder $\mathcal{E}$, more details in \cite{taming}. We also use the $L_2$ regularization on the codebook embeddings during quantization as in \cite{vitvqgan} and exponential moving average (EMA) for updating the codebook \cite{vq-vae} as they provide more stable training for the quantization block $\mathcal{Q}$. 

\textbf{Benefits of DSL} The advantages of learnable $\sigma_i$ in DSL are shown in Figure \ref{fig:method_freq} column 5 where the reconstructed activations show no checkerboard artifacts as in column 4. Moreover, the reconstructed activations $C_1$ are more similar to the encoder activations $A_1$ on the frequency spectrum. Compared to the baseline model (column 2), our model exhibits a more harmonious balance between low- and high-frequencies, leading to more accurate reconstructed images.

\subsection{Cross-attention Autoregressive Transformer (CAT)}

We further extend FA-VAE to the text-to-image generation task and introduce a new \textit{Cross-attention Autoregressive Transformer (CAT)}, presented in Figure \ref{fig:prior}. CAT uses all the token embeddings of a textual description given by the pretrained text encoder of CLIP model \cite{clip}, while existing works mostly use one or partial textual embeddings \cite{clip-gen, dalle, vqganclip}. Furthermore, CAT uses a cross-attention mechanism with the text tokens to guide the image generation at each step. This more fine-grained text condition allows the generation to capture more precisely the relationships of semantic attributes between text and image.


\begin{equation}
    p(s | c) = \prod_i p(s_i | s_{<i}, c)
\end{equation}
Then the predicted indices $z'$ can be decoded to an image using FA-VAE's decoder $G$. The loss is to maximize the log-likelihood of the data representations,
\begin{equation}
    \mathcal{L}_{\text{CAT}} = \mathbb{E}_{x \sim p(x)} [- \text{log} p(s)]
\end{equation}
In this paper, the GPT2 model \cite{gpt2}, which is the default setting in \cite{taming, clip-gen}, is utilized as the backbone for CAT, and cross-attention mechanism inspired from \cite{attention-all-you-need} is applied with the GPT2 structure.



\section{Experiments}
\label{sec:experiments}

\subsection{Experimental Details}

The datasets used in this paper are: (1) Multi-Modal CelebA-HQ dataset \cite{tedigan} with 30,000 high-resolution celebrity face images and each image comes with ten captions; (2) Flickr-Faces-HQ (FFHQ) dataset \cite{ffhq} of 70,000 high-resolution face images; (3) ImageNet \cite{imagenet} that contains around 1 million images from 1000 categories. Experiments are performed on V-100 GPUs and results are reported on the validation sets unless specified otherwise. For a fair comparison with existing models, the resolutions used for training are $256 \times 256$ on all datasets. The details of training settings and hyperparameters are provided in the supplementary materials. 

\subsection{Image Reconstruction}

\begin{table}\centering
\footnotesize
\setlength{\tabcolsep}{0.5mm}
\begin{tabular}{l | c c c c}
\toprule
Model & Dataset & Codebook Size & $(h \times w)$ & rFID $\downarrow$ \\
\midrule
RQ-VAE \cite{rq-vae}    & FFHQ    & 2048 & $(8 \times 8)$ & 5.33 \\
\textbf{FA-VAE (Ours)} & FFHQ    & 2048 &  $(16 \times 16)$ & 4.98 \\
\midrule
VQ-VAE-2 \cite{vqvae-2} & ImageNet & 512    & \makecell{$(64 \times 64)$ \\ \& $(32 \times 32)$}     & $\sim$ 10 (train) \\
VQ-GAN \cite{latent-diffusion}    & ImageNet & 8192  & $(64 \times 64)$ & 1.06 \\
\textbf{FA-VAE (Ours)} & ImageNet & 8192 & $(64 \times 64)$  & \textbf{0.40} \\
\midrule
DALL-E \cite{dalle}     & ImageNet &  8192  & $(32 \times 32)$    & 32.01 \\
VQ-GAN \cite{taming}    & ImageNet & 16384  & $(16 \times 16)$ & 5.15 \\
VQ-GAN \cite{taming}    & ImageNet & 1024  & $(16 \times 16)$ & 7.94 \\
VQ-GAN \cite{rq-vae}    & ImageNet & 16384  & $(8 \times 8)$ & 17.95 \\
$\text{RQ-VAE}^{\dagger}$ \cite{vq-vae}    & ImageNet & 16384  & $(8 \times 8)$    & 10.77 \\
RQ-VAE* \cite{rq-vae}    & ImageNet & 16384  & $(8 \times 8)$    & 4.73 \\
\textbf{FA-VAE (Ours)} & ImageNet & 16384 & $(16 \times 16)$  & \textbf{4.60} \\
\bottomrule
\end{tabular}
\captionsetup{aboveskip=3pt}
\caption{Reconstruction results on the validation data of FFHQ \cite{ffhq} and ImageNet \cite{imagenet} respectively, more results in the supplement. $\dagger$ means depth 2, and $*$ means depth 4 from RQ-VAE \cite{rq-vae}.}
\vspace{-0.5cm}
\label{tab:reconstruction}
\end{table}

\begin{table*}\centering
\footnotesize
\setlength{\tabcolsep}{1.3mm}
\begin{tabular}{l | l | c c | c | c c c c c | c c c }
\toprule
& ablation on & $\text{FFL}(X, \hat {X})$ & \makecell{\text{SL}\\($\mathcal{A}_i, \mathcal{C}_i)$} & \makecell{FCM \\ $\mathcal{F}$} & \makecell{kernel \\ $K_i(\mu, \sigma_i)$}  & \makecell{\text{DSL}\\($\mathcal{A}_i, \mathcal{C}_i)$} & \makecell{kernel size \\ $\mu$} & \makecell{ initial value \\ $\sigma_i$} & \makecell{pair-wise \\ $\sigma_i$} & $\mathcal{L}_1$ $\downarrow$ & $\mathcal{L}_{pips}$ $\downarrow$ & rFID $\downarrow$ \\
\midrule
1 & VQ-GAN \cite{taming} & \gray{\xmark} & \gray{\xmark} & \gray{\xmark} & \gray{\xmark} & \gray{\xmark} & \gray{\xmark} & \gray{\xmark} & \gray{\xmark} & 0.121 & 0.30 & 10.12 \\
2 & VQ-GAN + Style \cite{vitvqgan}  & \gray{\xmark} & \gray{\xmark} & \gray{\xmark} & \gray{\xmark} & \gray{\xmark} & \gray{\xmark} & \gray{\xmark} & \gray{\xmark} & 0.085 & 0.23 & 11.90 \\
3 & VQ-GAN \cite{ffl}       & \cmark  & \gray{\xmark} & \gray{\xmark} & \gray{\xmark} & \gray{\xmark} & \gray{\xmark} & \gray{\xmark} & \gray{\xmark} &  0.114  &  0.35    &  30.65  \\
\midrule
4 & \multirow{2}{*}{SL w/o kernel} 
          & \gray{\cmark}         &   \xmark &   \textbf{CONV} & \gray{\xmark} & \gray{\xmark} & \gray{\xmark} & \gray{\xmark} & \gray{\xmark} &   0.082   & 0.22   & 7.04     \\
 5   & \multirow{2}{*}{}         & \gray{\cmark}        &   \cmark   & \textbf{CONV} & \gray{\xmark} & \gray{\xmark} & \gray{\xmark} & \gray{\xmark} & \gray{\xmark}  &  0.082  & 0.22  &  7.02     \\
\midrule
6 & SL w/ kernel & \gray{\cmark} & \gray{\cmark} & \gray{CONV} & \cmark & \gray{\xmark} & \textbf{9} & \textbf{3} & \gray{\xmark} & 0.080 & 0.22 & 7.39 \\
\midrule 
             7 & \multirow{3}{*}{FCM architecture} 
             & \gray{\cmark} & \gray{\cmark} & \textbf{CONV} & \gray{\cmark} & \cmark & \gray{9} & \gray{3} & \gray{\xmark} & 0.081    & 0.21  &  5.90  \\
             8 & \multirow{3}{*}{} & \gray{\cmark} & \gray{\cmark} & \textbf{RES} & \gray{\cmark} & \cmark & \gray{9} & \gray{3} & \gray{\xmark} & 0.078    & 0.21  &  6.44 \\
             9 & \multirow{3}{*}{} & \gray{\cmark} & \gray{\cmark} & \textbf{ATTN} & \gray{\cmark} & \cmark & \gray{9} & \gray{3} & \gray{\xmark} & 0.089    & 0.23  &  7.49 \\
 \midrule
10 & \multirow{5}{*}{DSL kernel size $\mu$} 
            & \gray{\cmark} & \gray{\cmark} & \gray{RES} & \gray{\cmark} & \gray{\cmark} & \textbf{3} & \gray{3} & \cmark & 0.081    & 0.21  &  6.66 \\
    11  & \multirow{5}{*}{}        & \gray{\cmark} & \gray{\cmark} & \gray{RES} & \gray{\cmark} & \gray{\cmark} & \textbf{5} & \gray{3} & \cmark & 0.081    & 0.21  &  7.04 \\
     12  & \multirow{5}{*}{}            & \gray{\cmark} & \gray{\cmark} & \gray{RES} & \gray{\cmark} & \gray{\cmark} & \textbf{9} & \gray{3} & \cmark & 0.082    & 0.22  &  6.53  \\
     13  & \multirow{5}{*}{}            & \gray{\cmark} & \gray{\cmark} & \gray{RES} & \gray{\cmark} & \gray{\cmark} & \textbf{11} & \gray{3} & \cmark & 0.083    & 0.22  &  7.48  \\
      14  & \multirow{5}{*}{}           & \gray{\cmark} & \gray{\cmark} & \gray{RES} & \gray{\cmark} & \gray{\cmark} & \textbf{15} & \gray{3} & \cmark & 0.083    & 0.22  &  6.36  \\
\bottomrule
\end{tabular}
\captionsetup{aboveskip=3pt}
\caption{Ablation studies on the CelebA-HQ validation dataset \cite{celeba-hq}, visualizations are in Figure \ref{fig:celeba-ablation}. Words put in bold to highlight the changing component.}
\vspace{-0.5cm}
\label{tab:celeba-ablation}
\end{table*}

\begin{figure}
    \centerline{\includegraphics[width=0.5\textwidth]{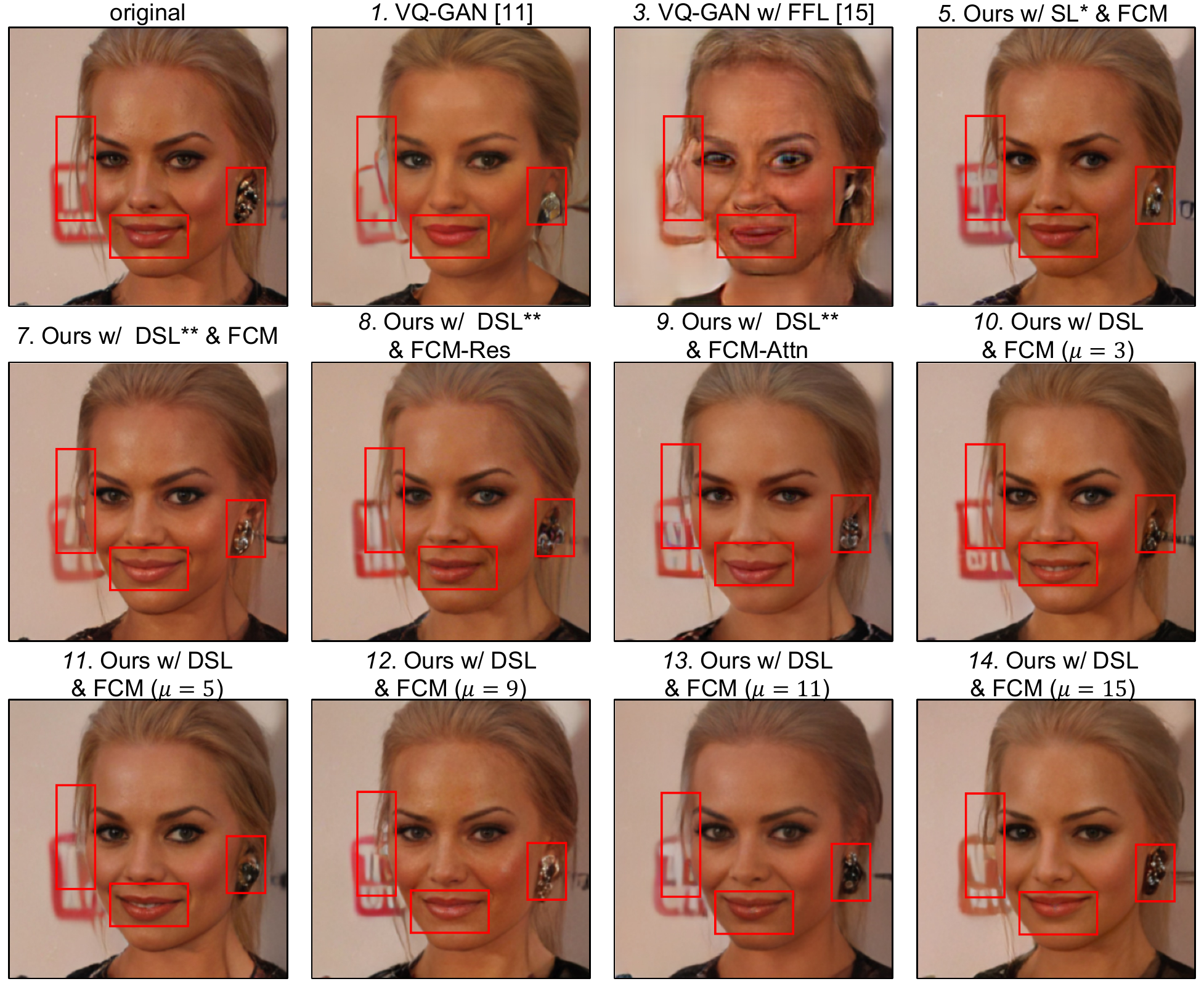}}
    \captionsetup{aboveskip=3pt}
    \caption{\small 
    \small{Reconstruction comparisons for ablation studies on CelebA-HQ. The figure number represents the setting in the corresponding row in Table \ref{tab:celeba-ablation}. $\text{SL}^*$ is SL without Gaussian kernel, $\text{DSL}^{**}$ is DSL with non pairwise $\sigma$. FCM is convolution architecture by default.}}
    \label{fig:celeba-ablation}
    \vspace{-0.6cm}
\end{figure}

\begin{figure}
    \centerline{\includegraphics[width=0.5\textwidth]{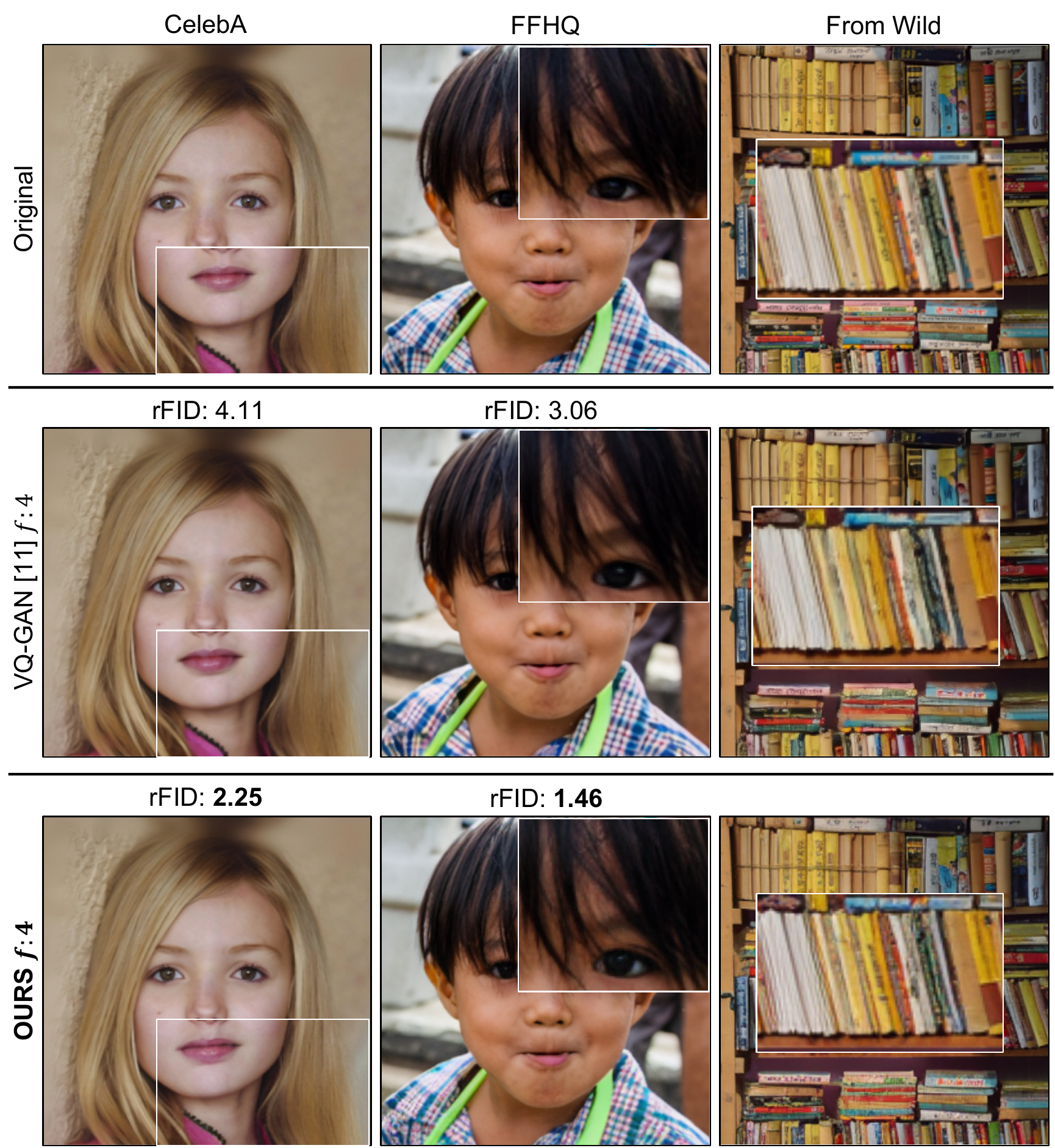}}
    \captionsetup{aboveskip=3pt}
    \caption{\small 
    \small{Zero-shot reconstruction on CelebA-HQ \cite{celeba-hq}, FFHQ \cite{ffhq} and photos from the wild, taken from \cite{taming}. The models are trained on ImageNet on different compression rates.}}
    \label{fig:zero-shot-vis}
    \vspace{-0.7cm}
\end{figure}

\begin{figure*}
    \centerline{\includegraphics[width=1.\textwidth]{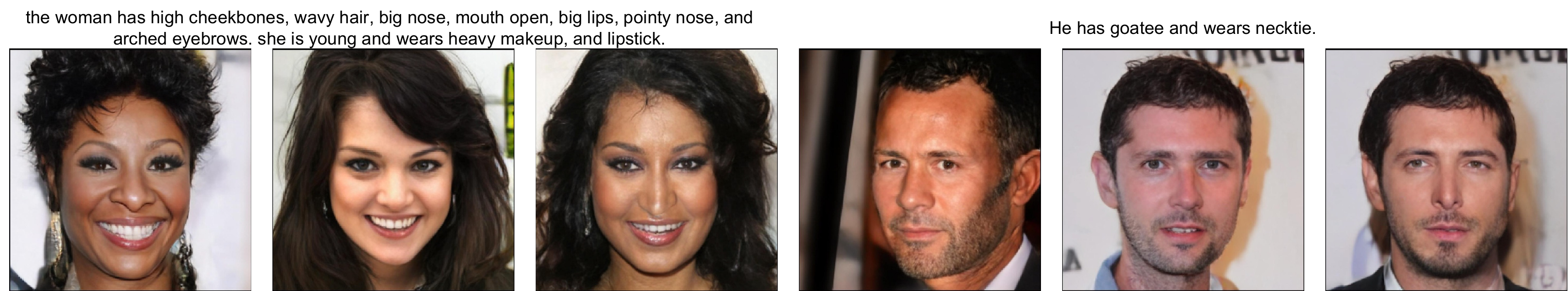}}
    \caption{\small 
    \small{Text-to-Image generation on CelebA-HQ-MM \cite{celeba-hq}}}
    \captionsetup{aboveskip=3pt}
    \label{fig:celeba-t2i}
    \vspace{-0.5cm}
\end{figure*}

\textbf{Reconstruction on FFHQ and Imagenet} The experiment results of reconstruction on FFHQ \cite{ffhq} and ImageNet \cite{imagenet} are presented in Table \ref{tab:reconstruction}. The results of the baselines are from the original paper. Note that smaller $h$ and $w$ means that the downsampling factor or the compression rate $f$ is larger. FA-VAE model shows improved reconstruction quality over baseline models across different compression rates in both datasets. The main reason is that FA-VAE can successfully reconstruct the important middle and high-frequencies which are neglected in baseline models. 

Figure \ref{fig:recon-imagenet} further supports our previous claim and demonstrates clear diverging reconstruction qualities between baseline models VQ-GAN \cite{taming}, DALL-E \cite{dalle}, and our FA-VAE model. With a larger compression rate, Figure \ref{fig:recon-imagenet} row 2 shows that more middle and high frequencies are compressed in VQ-GAN and DALL-E. In comparison, FA-VAE can reconstruct middle and high frequencies more accurately, which translates to good representations with improved semantics and local details in Figure \ref{fig:recon-imagenet} row 1. For instance, at a compression rate of 16, the dog's tongue of FA-VAE's reconstruction is more similar to the original image than VQ-GAN in terms of color and shape; more qualitative results are in the supplement.

\textbf{Ablation Studies on FCM and DSL} In Table \ref{tab:celeba-ablation}, we perform ablation studies on different architectures of FCM in combination with different settings of the Spectrum Loss (SL) and Dynamic Spectrum Loss (DSL). Figure \ref{fig:celeba-ablation} gives the accompanying visualizations of the quantitative results. 

First, as motivated in the Section \ref{sec:method}, Table \ref{tab:celeba-ablation} row 3 shows that considering all frequencies as equally important leads the VQ-GAN model to poor performance in terms of rFID and lpips (see Figure \ref{fig:celeba-ablation} image 3). In comparison, results in Table \ref{tab:celeba-ablation} rows 4 demonstrate that FCMs alone can help FA-VAE model achieve a better reconstruction because the residual connections in FCMs preserve better the information flow. The reconstruction is further improved when combined with a simple SL on images level (row 5) or blocks level (row 6). Although SL (row 6) shows slightly inferior reconstruction due to neglection of frequency importance variance across decoder blocks, Figure \ref{fig:celeba-ablation} image 5 shows that the details, such as earrings and hairs, can still be more enhanced than the baseline model shown in image 1. 

Then, Table \ref{tab:celeba-ablation} rows 7-9 compare the performance when the architecture of FCM varies as illustrated in Figure \ref{fig:fcm-architectures}; note that instead of using same $\sigma_i$ for $(\mathcal{A}_i, \mathcal{C}_i)$, we use two learnable $\sigma$ for $\mathcal{A}_i$ and $\mathcal{C}_i$ respectively. Quantitatively, when everything else is held equal, the convolution architecture shows better reconstruction than residual and attention architectures. Similarly, Figure \ref{fig:celeba-ablation} image 7 resembles the original image more than images 8 and 9 in terms of face radiance, smoothness, lip color, and shape. The reason is that residual connection enriches the outputs of FCM, which defies the purpose of enriching features of higher frequencies of the decoder features motivated in section \ref{sec:fcm}. The attention mechanism is at a disadvantage here because the decoder and the encoder exclusively use convolutions. 

Finally, varying the size of kernels in DSL (Table \ref{tab:celeba-ablation} row 10 - 14) show quite similar quantitative reconstruction performance, while qualitatively, Figure \ref{fig:celeba-ablation} shows that a larger kernel size tends to produce smoother reconstructions. One reason could be that a larger kernel size tends to smooth more the feature maps because more surrounding values are used during convolution. Thus, in other experiments, we choose a kernel size of 3, and the effects of kernel sizes are open to future works.

\textbf{Zero-shot Reconstruction} To further demonstrate the reconstruction capabilities of FA-VAE models, Table \ref{tab:zero_shot} gives the zero-shot reconstruction performance evaluated on CelebA-HQ \cite{celeba-hq} and FFHQ \cite{ffhq} using models trained on ImageNet \cite{imagenet}. The accompanying qualitative results are in Figure \ref{fig:zero-shot-vis}. Overall, FA-VAE displays impressive transferability capability with more faithful reconstruction in details, for instance, the light contrast in column 1 and book details in column 3. Note also that the common metric used in image compression PSNR is not perfectly correlated with rFID, but we put the metric here for reference.


\begin{table}[t!]\centering
\footnotesize
\setlength{\tabcolsep}{2.0mm}
\begin{tabular}{l |  c | c c | c c }
\toprule
\multicolumn{2}{c}{} & \multicolumn{2}{c}{CelebA} & \multicolumn{2}{c}{FFHQ}  \\
\midrule
Pretrained Model & $f$  & rFID $\downarrow$ & PSNR $\uparrow$ & rFID $\downarrow$ & PSNR $\uparrow$ \\
\midrule
VQ-GAN \cite{latent-diffusion}  & 16  & 8.62  & \textbf{23.40} & 6.83  & \textbf{22.68}\\
VQ-GAN \cite{taming}            & 4  & 4.11 & 31.20   & 3.06    & 30.82 \\
\midrule
\textbf{FA-VAE (Ours)} & 16  & \textbf{6.52} & 22.59     & \textbf{6.19} & 21.95 \\
\textbf{FA-VAE (Ours)}             & 4 &  \textbf{2.25} & \textbf{31.39}     & \textbf{1.46} & \textbf{30.85} \\
\bottomrule
\end{tabular}
\captionsetup{aboveskip=3pt}
\caption{Zero-shot reconstruction results on the validation data of FFHQ \cite{ffhq} and CelebA-HQ \cite{celeba-hq} using models trained on ImageNet \cite{imagenet}. $f$ is the downsampling factor, the codebook sizes for $f=\{16, 8, 4\}$ are $\{16384, 16384, 8192\}$ respectively. }
\vspace{-0.25cm}
\label{tab:zero_shot}
\end{table}

\subsection{Image Synthesis}


\begin{table}[t!]\centering
\footnotesize
\setlength{\tabcolsep}{4.0mm}
\begin{tabular}{l | c }
\toprule
Model & FID $\downarrow$ \\
\midrule
AttnGAN \cite{attngan}    & 125.98 \\
ControlGAN \cite{controlgan} & 116.32\\
DM-GAN \cite{dmgan}    &  131.05  \\
DF-GAN \cite{dfgan}    &  137.60 \\   
TediGAN \cite{tedigan} & 106.37  \\
LAFITE \cite{lafite}    & 12.54  \\
\midrule
\textbf{CAT (Ours)} & \textbf{10.23} \\
\bottomrule
\end{tabular}
\captionsetup{aboveskip=3pt}
\caption{Text-to-image generation on CelebA-HQ MM \cite{tedigan}.}
\vspace{-0.7cm}
\label{tab:generation-celeba}
\end{table}

Table \ref{tab:generation-celeba} shows the text-to-image generation performance on CelebA-HQ-MM \cite{celeba-hq}. Our proposed autoregressive transformer CAT yields a better generation quality than other GAN-based models, including AttnGAN \cite{attngan}, ControlGAN \cite{controlgan}, which are solely designed for image generation. Figure \ref{fig:celeba-t2i} shows that CAT can generate satisfactory images conditioned on text inputs of varying lengths. Attributes such as ``mouth open" and ``arched eyebrows" are captured during generation because all the tokens embeddings are used as a condition which gives more precise guidance. More quantitative and qualitative results on different datasets are in the supplement.

\section{Conclusion}
\label{sec:conclusion}

In this paper, we introduce the Frequency Augmented VAE (FA-VAE) model, which aims to improve reconstruction quality by bridging the gaps in the frequency domains between original and reconstructed images. New modules named Frequency Complement Modules (FCM) are crafted and guided under the new (Dynamic) Spectrum Loss ((D)SL) to learn to complement the reconstructed features of missing frequencies. A new Cross-attention Autoregressive Transformer (CAT) is proposed for achieving more precise textual-image alignment in the text-to-image generation task. FA-VAE shows improved reconstruction on various datasets compared to SOTA methods.

\section*{Acknowledgement}
This research is supported in part by the National Science Foundation under award No. 2040209.

\clearpage
{\small
\bibliographystyle{ieee_fullname}
\bibliography{camera_ready_paper}
}

\end{document}